# Batch Active Learning via Coordinated Matching

**Javad Azimi** AZIMI@EECS.OREGONSTATE.EDU
**Alan Fern** AFERN@EECS.OREGONSTATE.EDU
**Xiaoli Z. Fern** XFERN@EECS.OREGONSTATE.EDU
**Glencora Borradaile** GLENCORA@EECS.OREGONSTATE.EDU
Department of EECS, Oregon State University, Corvallis, OR, 97330 USA

**Brent Heeringa** HEERINGA@CS.WILLIAMS.EDU
Department of Computer Science, Williams College, Williamstown, MA 01267 USA

## Abstract

We propose a novel batch active learning method that leverages the availability of high-quality and efficient sequential active-learning policies by approximating their behavior when applied for $k$ steps. Specifically, our algorithm uses Monte-Carlo simulation to estimate the distribution of unlabeled examples selected by a sequential policy over $k$ steps. The algorithm then selects $k$ examples that best matches this distribution, leading to a combinatorial optimization problem that we term "bounded coordinated matching". While we show this problem is NP-hard, we give an efficient greedy solution, which inherits approximation bounds from supermodular minimization theory. Experiments on eight benchmark datasets show that the proposed approach is highly effective.

## 1. Introduction

In this paper, we consider active learning of classification functions. We are given an initial set of $m$ labeled examples $\mathcal{D}_l = \{(x_1, y_1), (x_2, y_2), \ldots, (x_m, y_m)\}$, where $y_i$ is the target label for input $x_i$. In addition, we are given a pool of $n$ unlabeled inputs $\mathcal{D}_u = \{x_1, x_2, \ldots, x_n\}$ for which the labels can be queried. The problem of *active learning* is to select the most informative examples (queries) from $\mathcal{D}_u$ to be labeled, so that the accuracy of the classifier increases quickly as the set of labeled examples grows. Active learning typically works in iterations, where each iteration builds a classifier based on the current training set, and then selects the examples to be labeled. The labeled examples will then be added to the training set and this procedure is repeated until we reach a good model or we exceed the labeling budget. Much existing active learning work has focused on a *sequential* instance of this framework where one example is selected to be labeled in each iteration. A number of sequential active learning methods have been developed that yield substantial empirical gains over their passive learning counterparts, including simple strategies such as the minimum margin and maximum uncertainty principles (Settles, 2009).

Batch active learning differs from sequential methods by selecting a batch of $k > 1$ examples to be labeled at each iteration (Hoi et al., 2006b; Brinker, 2003; Guo & Schuurmans, 2007). This batch-mode of active learning is often preferable to sequential methods when each label takes substantial time but can be produced in parallel. Such scenarios may arise when labels require running wet lab experiments, careful human analysis, or expensive computational processes.

A naive approach to batch active learning is to simply apply an existing sequential selection rule $k$ times to generate a batch, e.g. selecting the $k$ minimum margin examples. This approach, however, will often perform poorly since it will tend to ignore redundancy among selected examples. To address this issue, there has been a small amount of work on batch active learning, which provides different heuristic approaches for incorporating batch diversity into the selection method. For example, Brinker (2003) introduced an SVM-based batch approach, which selects a batch that minimizes the margin of the selected examples while maximizing their diversity. Hoi et al. (2006b; 2006a) chose a batch of examples that effectively maximizes the Fisher information of a classification model, which leads to a trade-off between uncertainty and diversity. Guo and Schuurmans (2007) posed batch active learning as a complex optimization problem that maximizes the discrimina-tive classification performance while taking into consideration the unlabeled examples. Unfortunately the resulting optimization problem is non-convex and requires heuristic fixes involving many parameters to work effectively. Recently, Guo (2010) introduced an approach that selects a batch of examples that maximizes the mutual information between unlabeled and labeled examples. The proposed combinatorial optimization problem is NP-hard and a heuristic algorithm is introduced to produce a solution.

This paper considers a general approach for batch active learning, which we refer to as "simulation matching". We are motivated by the observation that sequential methods are generally more example-efficient than their batch counterparts, since each example is selected with more information. Indeed in theory the best sequential strategy will never be worse than the best batch strategy, since one could simulate the batch approach and then select the examples sequentially. Leveraging the availability of highly-effective sequential active learning methods, we view a given sequential method as a gold-standard whose performance, in terms of label efficiency, we would like to approach using batch selection. That is, we aim to come close to the per-example accuracy improvement of the sequential method but in less time (fewer iterations) via batch selection.

For this purpose, we use Monte-Carlo simulation to estimate the posterior distribution over examples selected by the sequential method, and then select a batch of $k$ examples that "best matches" this distribution. A key contribution of our work is to instantiate the notion of "best match" by developing a novel matching objective called *bounded coordinated matching*. While we show that optimizing the objective is NP-Hard, we introduce an efficient greedy algorithm that optimizes the objective with an approximation bound. Our proposed algorithm is simple to implement and its scalability in terms of the number of unlabeled examples is similar to the base sequential policy. Experiments on eight benchmark datasets with different batch sizes demonstrate that the proposed approach is highly effective.

## 2. Simulation Matching for Batch Selection

Given a dataset $\mathcal{D}_l$ of labeled examples, we now consider how to select the next batch of $k$ examples to be labeled. A key issue in making this choice is to manage the tradeoff between selecting examples that individually look most informative for learning versus selecting a diverse set of examples. For instance, a common measure of informativeness is the margin, or class uncertainty, of an example with respect to the currently learned classifier. However, picking the top $k$ most informative examples under such measures will often select clusters of nearby examples that are quite redundant. Previous work on batch active learning has considered various objective functions for capturing this tradeoff and then searches for batches that approximately optimize those objectives.

In this work, we follow a different approach motivated by the fact that sequential active learning has been widely studied and a variety of computationally efficient and empirically effective sequential policies exist. For example, selecting the example with highest class uncertainty is often a simple yet highly effective baseline approach. The main idea behind our batch approach is to leverage such sequential policies by selecting a batch of $k > 1$ samples that "closely matches" the sequential policy's expected behavior. This idea has been explored recently for the very different problem of batch Bayesian optimization (Azimi et al., 2011). In that work, it was also the case that good sequential policies were available. However, since that work was focused on function optimization, the notion of "closely matching" used there is not suitable for active learning. The main contribution of our work is to develop a principled adaptation of the approach to batch active learning and to demonstrate its effectiveness.

### 2.1. Sequential Policy Simulation

Let $\pi$ be a sequential active learning policy. Given a set of labeled examples $\mathcal{D}_l$ and unlabeled examples $\mathcal{D}_u$, $\pi$ returns the next example $x \in \mathcal{D}_u$ to be labeled. We would like to "closely match" the behavior of $\pi$ when applied for $k$ steps. However, without the labels of the selected instance, we do not know how $\pi$ would behave after the first example is selected. In particular, different label outcomes will likely lead $\pi$ to select a different set of $k$ examples. In this work, we assume the availability of a posterior distribution of the labels of any example $x$ given $\mathcal{D}_l$, which can be estimated using a probabilistic classifier. A $k$-step executions of $\pi$ will result in a set of $k$ selected examples from $\mathcal{D}_u$. Let $S_\pi^k$ be the random variable denoting the set of $k$ examples resulting from such a $k$-step execution of $\pi$, which has a well defined distribution $P_\pi^k(\cdot)$ over the subsets of $\mathcal{D}_u$.

Importantly, it is generally straightforward to use Monte-Carlo simulation to draw samples of $S_\pi^k$. For example, this can be done by starting with $\mathcal{D}_l$ and selecting the first example $x_1$ using $\pi$. Then, we realize $y_1$, the class label of $x_1$, by sampling from the label posterior distribution of $x_1$. This simulated labeled example is then added to $\mathcal{D}_l$ and the process repeats for $k-1$ additional iterations to obtain a total of $k$ examples. Our batch policy is based on generating a number of samples of $S_\pi^k$, which are used to define an objective for optimizing a batch of $k$ experiments. Below we derive this objective and describe its optimization.

### 2.2. Coordinated Matching Objective

Our goal is to select a batch $B$ of $k$ unlabeled examples that best "matches the behavior" of a base sequential policy $\pi$

conditioned on the currently labeled examples. More precisely, we consider a batch $B$ to be a good match if it has high probability (relative to other sets) under the dataset distribution $P_\pi^k$. Unfortunately, for all but trivial sequential policies $\pi$, there will be no closed form for $P_\pi^k$, which makes it challenging to directly optimize the probability of $B$. Thus, our approach is to first approximate $P_\pi^k$ via a simpler distribution $Q^k$ that captures essential aspects of $P_\pi^k$ and then return the batch $B$ that is optimized under $Q^k$.

**Matching Mixture Model.** One naive choice for $Q^k$ would be to represent it as a latent mixture model, e.g. a Gaussian mixture model, from which $k$ i.i.d. examples are drawn in order to produce a batch. While estimating such a model based on samples of $S_\pi^k$ would be relatively straightforward, e.g. via an EM algorithm, it would generally produce poor results. In particular, the i.i.d. nature of the model would typically assign high probability to batches containing redundant examples arising from the most probable Gaussian component. This fails to capture the highly dependent nature of examples in $S_\pi^k$, which will typically avoid such redundancy.

In order to partially capture the dependencies in $S_\pi^k$, we use a variant of the Gaussian Mixture Model (GMM), which we call the *k-Matching Mixture Model (k-MMM)*. Similar to GMMs, a $k$-MMM model consists of a set of $k$ $n$-dimensional Gaussian with mean vectors $\boldsymbol{\mu} = \{\mu_1, \ldots, \mu_k\}$ and covariance matrices $\boldsymbol{\Sigma} = \{\Sigma_1, \ldots, \Sigma_k\}$. In contrast to the i.i.d generative process assumed by GMMs, a $k$-MMM generates a set of $k$ points by sampling one point from each of the $k$ components.

Given a $k$-MMM model and a set of $k$ points $S = \{x_1, \ldots, x_k\}$, there are $k!$ possible ways that $S$ can be generated, each corresponding to one possible matching of the $k$ points to the $k$ components. Given such a matching $m$, let $m(i)$ denote the index of the model component that is matched to point $x_i$ and let $\mathcal{M}$ denote the set of all possible matchings. Assuming a uniform prior over possible matchings, the probability of observing $S$ given a $k$-MMM $Q^k$ can be written as:

$$
\begin{aligned}
Q^k(S) &= \sum_{m \in \mathcal{M}} Q^k(S, m) \\
&= \frac{1}{k!} \sum_{m \in \mathcal{M}} Q^k(S \mid m) \\
&= \frac{1}{k!} \sum_{m \in \mathcal{M}} \prod_{i=1}^{k} f(x_i; \mu_{m(i)}, \Sigma_{m(i)}),
\end{aligned} \tag{1}
$$

where $f$ is the Gaussian PDF.

Importantly, unlike an i.i.d. model, the point sets generated by $Q^k$ can be highly dependent since there is a strict requirement that each component generates exactly a single point. In our application, this is useful in that it can capture distributions that assign higher probability to diverse datasets, which is a typical characteristic of $P_\pi^k$.

**Estimating $Q^k$.** We now wish to select a $k$-MMM model $Q^k$ that best approximates our target distribution $P_\pi^k$. To simplify this estimation problem, in this work, we limit our attention to models where the means are selected from the unlabeled examples $\mathcal{D}_u$ and all of the $\Sigma_i$ are equal to a known $\Sigma$[1]. Under these assumptions, we can view the problem of selecting $Q^k$ as a combinatorial problem of selecting the best subset of $k$ points $\boldsymbol{\mu} = \{\mu_1, \ldots, \mu_k\}$ from $\mathcal{D}_u$ to serve as the component means. We will let $Q_\mu^k$ denote our model for a particular set $\boldsymbol{\mu}$. Our optimization objective is now to find the set $\boldsymbol{\mu}$ that minimizes the KL-divergence $\mathrm{KL}(P_\pi^k \| Q_{\boldsymbol{\mu}}^k)$, which is equivalent to minimizing the cross-entropy between the distributions given by $\mathrm{H}(P_\pi^k, Q_{\boldsymbol{\mu}}^k) = \mathbb{E}\left[-\log Q_{\boldsymbol{\mu}}^k(S_\pi^k)\right]$.

The resulting minimization problem is intractable due to the complicated nature of $P_\pi^k$. However, we can sample from this distribution using simulation as described previously and generate a set of samples $\mathcal{S} = \{S_1, \ldots, S_N\}$, which can be used to approximate the expectation. Let $U^k$ be the set of all size-$k$ subsets of unlabeled examples in $\mathcal{D}_u$, our optimization objective can be formulated as follows.

$$
\begin{aligned}
\arg\min_{\boldsymbol{\mu} \in U^k} \mathrm{H}(P_\pi^k, Q_{\boldsymbol{\mu}}^k) &\approx \arg\max_{\boldsymbol{\mu} \in U^k} \sum_{i=1}^{N} \log Q_{\boldsymbol{\mu}}^k(S_i) \\
&= \arg\max_{\boldsymbol{\mu} \in U^k} \sum_{i=1}^{N} \log \sum_{m \in \mathcal{M}} Q_{\boldsymbol{\mu}}^k(S_i, m)
\end{aligned}
$$

To further simplify the above objective, we note that for the purpose of maximizing over the means $\boldsymbol{\mu}$, $\sum_{m \in \mathcal{M}} Q_{\boldsymbol{\mu}}^k(S_i, m)$ can be reasonably approximated by $\max_{m \in \mathcal{M}} Q_{\boldsymbol{\mu}}^k(S_i, m)$. This is because the value of $Q_{\boldsymbol{\mu}}^k(S_i, m)$ decays very quickly for non-optimal matchings, since such matchings will typically assign a data point in $S_i$ to a component with a distant mean. Thus, our objective is approximated by:

$$
\begin{aligned}
&\arg\max_{\boldsymbol{\mu} \in U^k} \sum_{i=1}^{N} \log \max_{m \in \mathcal{M}} Q_{\boldsymbol{\mu}}^k(S_i, m) \\
&= \arg\max_{\boldsymbol{\mu} \in U^k} \sum_{i=1}^{N} \max_{m \in \mathcal{M}} \log Q_{\boldsymbol{\mu}}^k(S_i, m) \\
&= \arg\min_{\boldsymbol{\mu} \in U^k} \sum_{i=1}^{N} \min_{m \in \mathcal{M}} \sum_{j=1}^{k} d_\Sigma(x_{ij}, \mu_{m(j)})
\end{aligned} \tag{2}
$$

where $x_{ij}$ is the $j$-th example in set $S_i$, and the distance

[1]Generally speaking, the values of the $\Sigma_i$ may be set within a cross validation process, or via standard heuristic rules.

**Algorithm 1** Greedy Supermodular Minimization Algorithm

**Require:** Set function $g$, Finite set $\mathcal{A}$.
**Ensure:** $\mu \subseteq \mathcal{A}$ such that $|\boldsymbol{\mu}| = k$

```
μ ← A
while |μ| > k do
    x ← arg min_{x∈μ} g(μ \ x)
    μ ← μ \ x
end while
return μ
```

$d_\Sigma(x_1, x_2) = (x_1 - x_2)'\Sigma^{-1}(x_1 - x_2)$ can be interpreted as the cost of matching $x_1$ to $x_2$. In our experiments, we use the identity matrix for $\Sigma$, resulting in Euclidean distance.

The above objective corresponds to a novel optimization problem that we call *Bounded Coordinated Matching (BCM)*. Given a particular choice of $\boldsymbol{\mu}$, we find the minimum cost matching between $\boldsymbol{\mu}$ and each set $S_i \in \mathcal{S}$. The overall cost of $\boldsymbol{\mu}$ is the sum of all $N$ costs, i.e. $\sum_{i=1}^N \min_{m\in\mathcal{M}} \sum_{j=1}^k d_\Sigma(x_{ij}, \mu_{m(j)})$ . The bounded coordinated matching problem involves finding the $\boldsymbol{\mu}$ that achieves minimum overall cost. Since minimum cost matchings between two sets can be found in polynomial time via the Hungarian algorithm, the overall cost for any $\boldsymbol{\mu}$ can be computed efficiently. Unfortunately, the problem of finding the optimizing set $\boldsymbol{\mu}$ is NP-complete.

**Theorem 1.** *BCM is NP-complete (see appendix for the proof).*

Fortunately, BCM allows for certain approximation guarantees to be made for a simple greedy algorithm, which we present in Section 3.

**Summary of Approach.** To summarize, our overall approach is as follows. First, we use simulation to generate $N$ independent sample trajectories $S_1, \cdots, S_N$ of a sequential active learning policy (we use the maximum entropy policy in our experiments). Second, we approximate the distribution generating these trajectories by $Q^k_{\boldsymbol{\mu}}$ where the set $\boldsymbol{\mu}$ is found by approximately optimizing Objective (2) (see next section for the optimization approach). Finally, given $Q^k_{\boldsymbol{\mu}}$ we select a batch of $k$ unlabeled data points $B$ such that $B = \arg\max_B \max_{m\in\mathcal{M}} Q^k_{\boldsymbol{\mu}}(B, m)$. Since we limited the choice of $\boldsymbol{\mu}$ to subsets of the unlabeled data, it is easily verified that $B$ is simply equal to $\boldsymbol{\mu}$, which is returned as the batch of examples for which to request labels.

## 3. Optimization Approach

In this section we present a greedy approximation algorithm for BCM motivated by theoretical results on the minimization of non-increasing, supermodular set functions.

### 3.1. Greedy Approximation Algorithm

**Definition 1.** *Given a finite set $\mathcal{A}$, a function on subsets of $\mathcal{A}$, $g : 2^{\mathcal{A}} \to \mathbb{R}^+$ is supermodular if for all $A_1 \subseteq A_2 \subseteq \mathcal{A}$ and $x \in \mathcal{A}\setminus A_2$, it holds that $g(A_1) - g(A_1 \cup x) \geq g(A_2) - g(A_2 \cup x)$.*

In other words, a supermodular function demonstrates "diminishing returns" because adding an element to set $A \subseteq \mathcal{A}$ decreases the value of $g(\cdot)$ by at most as much as adding the element to a subset of $A$. In addition, a set function is non-increasing if for any set $A$ and element $x$ we have $g(A) \geq g(A \cup \{x\})$. It turns out that the problem of finding a size $k$ subset of $\mathcal{A}$ that minimizes a non-increasing supermodular function $g(\cdot)$ can be approximately solved via a simple greedy algorithm. Algorithm 1 outlines this approach, which simply starts with all elements of $\mathcal{A}$ and iteratively removes the element whose removal leads to the smallest increase in $g(\cdot)$ until only $k$ elements remain. We have the following known guarantee.

**Theorem 2.** ***(Il'ev, 2001)*** *Let $g(\cdot)$ be a monotonic non-increasing supermodular function over subsets of the finite set $\mathcal{A}$, $|\mathcal{A}| = m$ and $g(\mathcal{A}) = 0$. Let $\boldsymbol{\mu}$ be the set of the elements returned by the greedy algorithm 1 s.t $|\boldsymbol{\mu}| = k$, $q = m - k$ and $\boldsymbol{\mu}^* = \mathrm{argmin}_{\boldsymbol{\mu}'\subseteq\mathcal{A},|\boldsymbol{\mu}'|=k}\, g(\boldsymbol{\mu}')$, then*

$$g(\boldsymbol{\mu}) \leq \frac{1}{t}\left[\left(\frac{q+t}{q}\right)^q - 1\right] g(\boldsymbol{\mu}^*) \leq \frac{e^t - 1}{t} g\left(\boldsymbol{\mu}^*\right) \quad (3)$$

*where t is the steepness parameter of function $g(\cdot)$ which is defined as:*

$$\begin{aligned} t &= \frac{s}{s-1} \quad s.t. \\ s &= \max_{x\in\mathcal{A}} \frac{(g(\emptyset) - g(x)) - (g(\mathcal{A}\setminus x) - g(\mathcal{A}))}{g(\emptyset) - g(x)} \end{aligned} \quad (4)$$

Notice that the approximation bound involves the steepness parameter $t$ of $g(\cdot)$, which characterizes the rate of decrease of $g(\cdot)$. This is unavoidable because achieving a constant factor approximation guarantee is not possible unless P=NP (Nemhauser & Wolsey, 1999). Furthermore, this bound has been shown to be tight for any $t$ (Il'ev, 2001). Note that this is in contrast to guarantees for greedy maximization of submodular functions (G. L. Nemhauser & Fisher, 1978) for which there are constant factor guarantees. In addition, the greedy algorithm we use is qualitatively different from the one used for submodular maximization, since it greedily removes elements from $\boldsymbol{\mu}$ rather than greedily adding elements to $\boldsymbol{\mu}$.

The objective function corresponding to the BCM optimization problem (2) is the following function over subsets $\boldsymbol{\mu}$ of the unlabeled data points $\mathcal{D}_u$:

$$g(\boldsymbol{\mu}) = \sum_{i=1}^{N} \min_{m\in\mathcal{M}} \sum_{j=1}^{k} d_\Sigma(x_{ij}, \mu_{m(j)}) \quad (5)$$

The BCM problem corresponds to minimizing this function subject to $|\boldsymbol{\mu}| = k$. It is easily verified that this objective is a non-increasing supermodular function of $\boldsymbol{\mu}$. Further, since the points $x_{ij}$ in the objective are elements of $\mathcal{D}_u$, we have that $g(\mathcal{D}_u) = 0$. Therefore, $g(\cdot)$ satisfies all of the properties for Theorem 2 and the greedy algorithm provides the corresponding guarantee. Thus, we use the above greedy algorithm applied to the function $g(\cdot)$ and set $\mathcal{D}_u$ as our BCM optimizer, i.e. $A = \mathcal{D}_u$ in Algorithm 1.

### 3.2. Accelerated Greedy Algorithm

Each iteration of the greedy algorithm requires evaluating the cost function (Equation 5) for removing each element $x$ from the current set $\boldsymbol{\mu}$, which is at most the size of $\mathcal{D}_u$. Each cost function evaluation involves finding $N$ minimum cost matchings, between each of the $S_i$ and $\boldsymbol{\mu} \setminus x$ (when $\boldsymbol{\mu} \setminus x$ is larger than $S_i$, some elements of $\boldsymbol{\mu}$ are unmatched), which can be done via $N$ calls to the Hungarian algorithm. While polynomial, for a naive implementation, each iteration can be computationally expensive when $\mathcal{D}_u$ is large. Fortunately, there are at least three ways to soundly speedup the computation, leading to drastic time reductions in our experience and allowing the computation to be independent of the size of $\mathcal{D}_u$.

First, let $\boldsymbol{\mu}$ be the current set and $\boldsymbol{\mu}_i \subseteq \boldsymbol{\mu}$ be the set of elements in $\boldsymbol{\mu}$ that are matched to $S_i$ in the minimum matching. It is easy to verify that $g(\boldsymbol{\mu}) = g(\cup_i \boldsymbol{\mu}_i)$. This observation implies that instead of initializing $\boldsymbol{\mu}$ to be the entire unlabeled data set $\mathcal{D}_u$, we can soundly initialize $\boldsymbol{\mu}$ to be $\mu_0 = \cup_i S_i$ since the minimum matching between $S_i$ and $\mathcal{D}_u$ must be $S_i$ itself. Thus, the time complexity of the greedy algorithm under this initialization grows with the size of $\mu_0$ (the number of data points generated during simulation, which is at most $N \cdot k$), rather than the potentially much larger $\mathcal{D}_u$. In other words, the run time of the greedy algorithm is independent of the size of $\mathcal{D}_u$, which is often quite large.

Second, for points $x$ that are unpruned by the first rule, we can often avoid computing $g(\boldsymbol{\mu} \setminus x)$ by exploiting the supermodularity property. This idea is analogous to a similar speedup approach used for submodular maximization (Krause et al., 2008). From Definition 1, we can directly conclude that for $A_1 \subseteq A_2 \subseteq \mathcal{A}$, $g(A_1 \setminus x) - g(A_1) \geq g(A_2 \setminus x) - g(A_2)$. We define the non-negative *incremental difference* of an instance $x$ with respect to a set $\boldsymbol{\mu}$ to be $\delta(\boldsymbol{\mu}, x) = g(\boldsymbol{\mu} \setminus x) - g(\boldsymbol{\mu})$, which is the amount of increase of our objective function after removing a sample $x$ from $\boldsymbol{\mu}$. Normally this incremental difference must be computed for all $x \in \boldsymbol{\mu}$ in each iteration. However, by maintaining these incremental differences, we can often soundly avoid recomputing a large majority of them in any given iteration. The first iteration must compute the differences for all points in $\boldsymbol{\mu}$. We then sort the points in increasing order based on their incremental differences, and remove the first point. For the following iteration, we move on to the next point in the sorted list and recompute its incremental difference. If the value is still smaller than the remaining points, we can immediately remove this point from $\boldsymbol{\mu}$ and proceed to the next iteration without recomputing any other differences. Otherwise, we proceed to evaluate the next points in the sorted list until finding one whose recomputed difference is less than the other stored differences and remove the point. The supermodular property guarantees that this approach makes the same choices as the full greedy algorithm, but effectively avoids a large number of difference computations in practice.

Finally, for any point $x$ that we need to compute its difference and hence evaluate $g(\boldsymbol{\mu} \setminus x)$, we can reduce the cost of this computation by storing the set of maximum matchings between $\boldsymbol{\mu}$ and the $S_i$. In particular, rather than recomputing the maximum matching between $\boldsymbol{\mu} \setminus x$ and each $S_i$ from scratch, we can start with the current matching to $S_i$. If the current matching does not involve $x$, then no recomputing is needed. Otherwise the matching can be updated with a single shortest path computation. This results in a reduction in time complexity at least by a factor of $k$ compared to running the full Hungarian algorithm for each $S_i$. Details are described in the supplementary material.

## 4. Scalability

The computation of our batch selection approach can be divided into two stages: 1) Simulation of the sequential policy, and 2) Solving the resulting BCM problem. As the number of unlabeled data points $n$ increases, the simulation time will also increase, since each simulation step involves applying a sequential policy, which typically considers each unlabeled point. For typical sequential policies, including the one in our experiments, the time complexity will grow linearly in $n$. Fortunately, the $N$ simulations generated during the first stage are independent, which allows for easy parallelization, possibly resulting in a time reduction of a factor of $N$. That is, with parallelization there need not be any time overhead compared to running a typical sequential algorithm for $k$ steps.

Further, as described previously the time complexity of the second stage does not depend on $n$, but rather on $N \cdot k$. Overall, the scalability of the combined two stages in terms of $n$ is similar to the underlying sequential base policy.

## 5. Experimental Results

**Datasets.** In this section we evaluate our proposed batch active learning method using eight binary classification problems from the UCI machine learning repository (Asuncion & Newman, 2010) including (the number of sam-ples and attributes are shown in the parenthesis for each dataset): Breast(569, 32), Ionosphere(351, 34), Pima(768, 8), German(1000, 24), Haberman(306, 3), Sonar(208, 60), EF(1543, 16) and MN(1575, 16). The EF and MN datasets are subsets of the original multi-class letter dataset, created by retaining only letters E and F, M and N respectively.

**Baseline Algorithms.** To evaluate the efficacy of the proposed algorithm, we compare our algorithm against four baseline methods. For the first baseline, we consider the *Fisher Information* approach by Hoi et al.(Hoi et al., 2006b), one of the state-of-the-art methods in batch active learning. This method selects a batch that maximizes the Fisher information of a classification model (we use Kernel logistic regression in our experiments). The second baseline, which we call *Maximum Uncertain*, simply selects the top $k$ most uncertain (as measured by class entropy) examples to form the batch. This simple batch algorithm is a commonly used baseline in batch active learning literatures. We further include the "random" policy in our comparison, which has demonstrated very competitive performance in prior batch active learning studies (Guo & Schuurmans, 2007). Finally, we also compare to the sequential policy that selects the example with the highest class entropy, which is also the base sequential method that our simulation matching algorithm tries to match. Note that we also compared our algorithm against SVM-D, which is another batch selection algorithm based on the minimum margin principle (Brinker, 2003). The results were not reported here since it was consistently worse than other baseline batch active learning approaches. Note that we were not able to compare results to two recent batch active learning methods (Guo & Schuurmans, 2007), and (Guo, 2010) because we were not able to acquire a working implementation of these algorithms, and both algorithms involve complex optimization procedures that are non-trivial to implement and tune.

**Experimental Setting.** We use kernel logistic regression with an RBF kernel (kernel width= $0.05$) as our classifier for all algorithms. We use N=20 simulated trajectories for each batch selection and consider batch sizes of 10 and 20. Each dataset is randomly divided into 70% training data and 30% testing data. Active learning is initialize with five random examples per class from *train* and iteratively selects batches of unlabeled examples in *train* to query. The classification accuracy is evaluated after each batch selection on the test data. The entire process is repeated for 50 independent runs and the average results are reported.

**Evaluation.** We show the classification accuracy of different methods on the eight datasets in Figure 1(the batch size is shown in the parenthesis for each dataset). The $x-$axis indicates the number of queries and the $y-$axis represents the classification accuracy. First, let us focus on comparing our proposed method with the baseline batch methods including *Fisher Information* and *Maximum Uncertain*. We observe that, for most datasets, the learning curves produced by our method dominate the learning curves of the other batch methods. This is true for both batch sizes, but the improvements our method achieves is more significant and consistent for larger batch size of 20. Interestingly, we observe that *Fisher Information* and *Maximum Uncertain* results are sometimes dominated by *random*. For example, this is the case for Pima and German, where both methods performed consistently worse than *random* for both batch sizes. While surprising, this is actually consistent with what have been observed in a previous batch active learning study (Guo & Schuurmans, 2007). This suggests that it is actually non-trivial to design a batch active learning method that performs competitively to *random* in a consistent fashion. Notably, our proposed approach is the only method in our comparison that demonstrated this robustness, which is a highly desirable property of our method. In addition, the variance of our approach is quite small in our experiments. This is likely due to the average effect of using a set of simulations, which are processed in aggregate by the greedy optimizer. Among all data sets, Breast data set has the highest variance which is $0.006$ and $0.0074$ for batch size 10 and 20 respectively.

We also observe that the sequential method generally outperforms the batch methods (with a few exceptions) and more significantly for batch size 20. This is aligned with our expectation because a sequential approach makes more efficient use of the labeled examples when making selection choices. Interestingly, our proposed method is able to match the performance of sequential method reasonably well for many datasets, and even sometimes outperform sequential (e.g., Ionosphere and Breast). We conjecture this is due to the fact that our proposed method aggregates the outcome of many simulations, which may reduce the variance of the base sequential active learning procedure.

**Computational Time.** We compute the CPU run time of selecting a batch of 20 examples in one of our largest data sets, MN, on a standard desktop computer with 2.13 GHz CPU (dual core) and 2 GB of memory. It takes less than 3 minutes using an un-optimized Matlab implementation. This time is reasonable for most applications of batch active learning, where labeling time is generally significant. As discussed previously, this time can be reduced via parallelization and will grow reasonably with the size of $\mathcal{D}_u$.

## 6. Conclusions

In this paper, we introduce a novel method for batch active learning, which follows a recently proposed general approach named "simulation matching". The basic idea behind simulation matching is to design batches of queries by imitating the behavior of a high-quality sequential policy via simulation. While this general approach has been successfully applied to the problem of batch Bayesian optimization, the notion of "matching" used in prior work is not suitable for active learning. We put forth a principled adaptation of the simulation matching approach to batch active learning. In essence, we consider $S_\pi^k$, the set of $k$ points selected by sequential policy $\pi$, to be a random variable. Because the distribution of $S_\pi^k$ is too complex to directly estimate, we draw samples from this distribution via simulation and approximate the distribution using a $k$-Matching Mixture Model, which is then used to select the batch. This results in a combinatorial optimization problem that we call "bounded coordinated matching" (BCM), and we present an efficient algorithm that provides approximation guarantees. We evaluate the proposed approach on eight UCI datasets and the results show that our method is highly competitive compared to baseline methods.

## A. Proof of Theorem 1

*Proof.* Given $S_1, \ldots, S_N$, where each $S_i$ contains a set of $k$ points $\{x_{ij}\}_{j=1}^k$, the BCM objective function is:

$$\underset{\boldsymbol{\mu} \subset \mathcal{D}_u : |\boldsymbol{\mu}| = k}{\operatorname{argmin}} \sum_{i=1}^{N} \min_{m \in \mathcal{M}} \sum_{j=1}^{k} d_\Sigma(x_{ij}, \mu_{m(j)}) \qquad (6)$$

Consider the following graph representation of the problem. We define a weighted bipartite $G = (U, V, E)$ where $V = \{S_1, S_2, \cdots S_N\}$, and $U = \mathcal{D}_u$ representing the set of unlabeled examples. $E \subseteq U \times V$ is an edge set where the weight $w(u, v) = d_\Sigma(u, v)$. A *coordinated matching* on $G$ is a subset of edges $E' \subseteq E$ such that, for $1 \leq i \leq t$, $(U \times S_i) \cap E'$ is a matching. The weight of a coordinated matching $E'$ is $w(E') = \sum_{(u,v) \in E'} w(u, v)$. Given $G$ and some integer $k \leq |U|$, the *bounded, coordinated matching problem* (BCM) asks for a minimum weight coordinated matching such that the edges in the matching are, in total, incident to at most $k$ vertices from $U$. Below we will turn this minimization problem into an equivalent maximization problem by replacing all the weights in $G$ such that $w'(u, v) = \lambda - w(u, v)$, where $\lambda$ is a large constant to ensure all positive $w'$ values. It is easy to see that the solution of this maximization problem of BCM is also the solution to the original minimization problem.

Focusing on the maximization problem, the decision version of the problem augments $G$ and $k$ with a weight $W$ and asks if there exists a bounded, coordinated matching with weight $W$. We show that the decision version of maximizing BCM is NP-hard. We reduce from the well-known *3-Dimensional Matching* (3DM) problem which, given a set $T \subset X \times Y \times Z$ where $X$, $Y$, and $Z$ are *disjoint* and an integer $k$, asks if there exists a subset $M \subseteq T$ of size at least $k$ such that for any two distinct sets $M_i, M_j \in M$, $M_i \cap M_j = \emptyset$. 3DM remains NP-hard even when $|X| = |Y| = |Z| = k$ (i.e. it becomes an *exact-cover* problem). The reduction is the natural one: create an unweighted, bipartite graph $G = (U, V, E)$ where $V = X \cup Y \cup Z$ and $U = \{u_1, \ldots, u_k\}$ is a set of $k$ nodes. For every $t_i \in T$ where $t_i = \{x_i, y_i, z_i\}$ add the edges $\{u_i, x_i\}, \{u_i, y_i\}, \{u_i, z_i\}$ to $E$. It's clear that $T$ has has a 3-dimensional matching of size $k$ if and only if $\langle G, k \rangle$ has a bounded, coordinated matching of weight $3k$. $\square$

## Acknowledgments

The authors acknowledge the support of the NSF under grants IIS-0812514, IIS-1055113, and IIS-0905678.

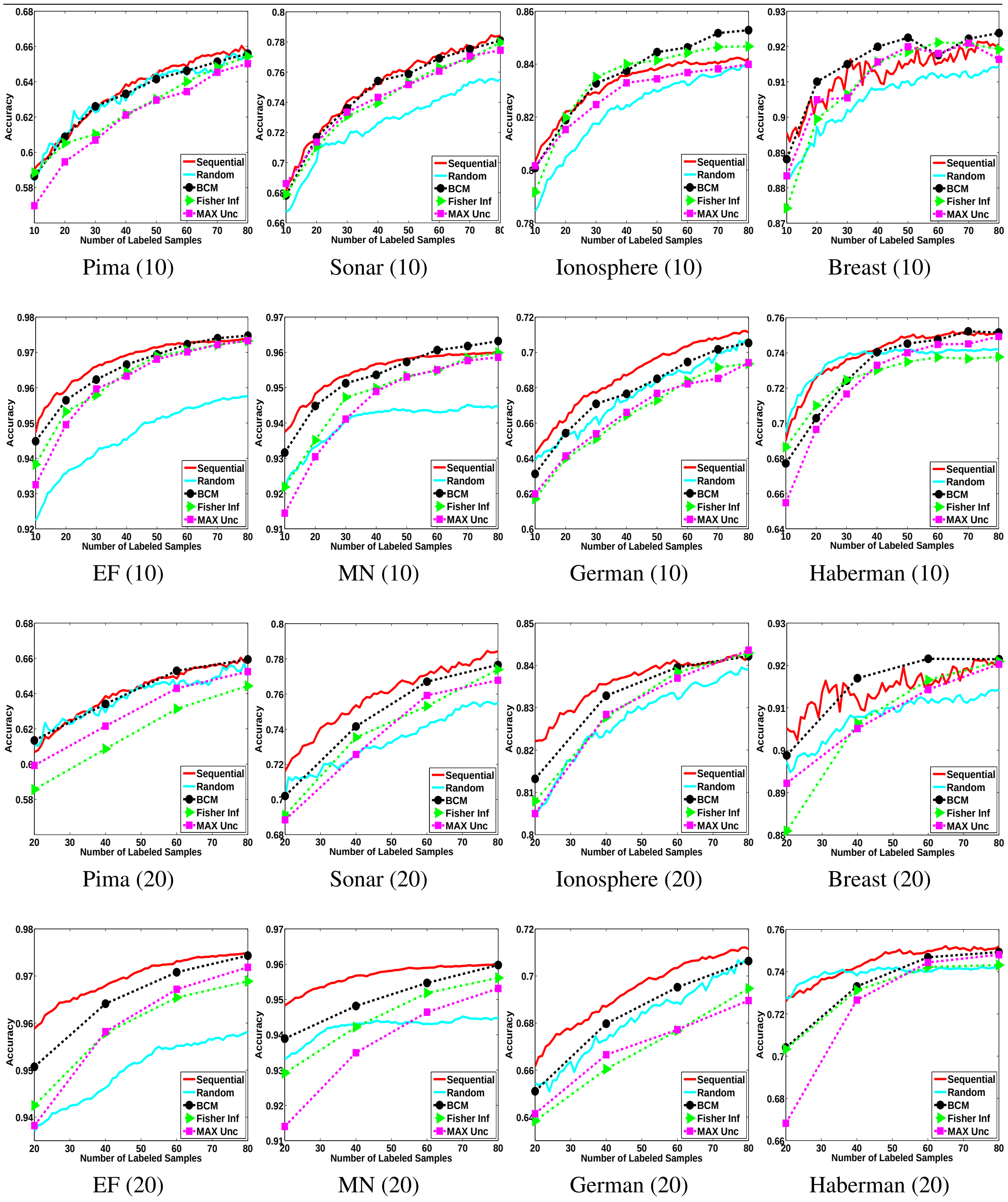

*Figure 1.* The performance of the proposed algorithm.